\documentclass[journal]{IEEEtran}

\ifCLASSINFOpdf
\else
   \usepackage[dvips]{graphicx}
\fi
\usepackage{url}

\usepackage{amsmath}
\usepackage{amsfonts}
\usepackage{graphicx}
\usepackage{CJKutf8}

\begin{document}

\title{VALLR-Pin: Uncertainty-Factorized Visual Speech Recognition for Mandarin with Pinyin Guidance}

\author{Chang Sun, Dongliang Xie, Wanpeng Xie, Bo Qin and Hong Yang
\thanks{
    C. Sun is with the State Key Laboratory of Networking and Switching Technology, Beijing University of Posts and Telecommunications, and the First Research Institute of the Ministry of Public Security of the People's Republic of China, Beijing, China. E-mail: sunch@bupt.edu.cn.
}
\thanks{
    \textbf{D. Xie (corresponding author)} and Wanpeng Xie are both with the State Key Laboratory of Networking and Switching Technology, Beijing University of Posts and Telecommunications, Beijing 100876, China.
}
\thanks{
    B. Qin and H. Yang are both with the First Research Institute of the Ministry of Public Security of the People's Republic of China, Beijing, China.
}}

\markboth{Journal of \LaTeX\ Class Files, Vol. 14, No. 8, August 2015}
{Shell \MakeLowercase{\textit{et al.}}: Bare Demo of IEEEtran.cls for IEEE Journals}
\maketitle

\begin{abstract}
Visual speech recognition (VSR) aims to transcribe spoken content from silent lip-motion videos and is particularly challenging in Mandarin due to severe viseme ambiguity and pervasive homophones. We propose VALLR-Pin, a two-stage Mandarin VSR framework that extends the VALLR architecture by explicitly incorporating Pinyin as an intermediate representation. In the first stage, a shared visual encoder feeds dual decoders that jointly predict Mandarin characters and their corresponding Pinyin sequences, encouraging more robust visual-linguistic representations. In the second stage, an LLM-based refinement module takes the predicted Pinyin sequence together with an N-best list of character hypotheses to resolve homophone-induced ambiguities. To further adapt the LLM to visual recognition errors, we fine-tune it on synthetic instruction data constructed from model-generated Pinyin-text pairs, enabling error-aware correction. Experiments on public Mandarin VSR benchmarks demonstrate that VALLR-Pin consistently improves transcription accuracy under multi-speaker conditions, highlighting the effectiveness of combining phonetic guidance with lightweight LLM refinement.
\end{abstract}


\IEEEpeerreviewmaketitle

\section{Introduction}
\label{sec:introduction}

\IEEEPARstart{V}{isual} Speech Recognition, also known as lip-reading, aims to transcribe spoken language solely from silent video recordings of lip movements. This capability has broad applications in hearing-impaired communication, silent speech interfaces, and speech recognition under adverse acoustic conditions. Despite recent progress driven by deep neural networks \cite{rezaee2025automatic} \cite{dixit2024cnn}, VSR remains a highly challenging task due to the inherent ambiguity of visual speech signals.

One fundamental difficulty of VSR lies in the \emph{viseme ambiguity} problem: where distinct phonemes may share nearly identical lip movements, while co-articulation further blurs temporal boundaries between adjacent units. In Mandarin, this challenge is further exacerbated by a mismatch between visual discriminability and linguistic representation, rendering Mandarin VSR inherently ill-posed. While visual cues can often constrain speech content at the syllable level, directly predicting characters from visual input requires resolving a one-to-many mapping, in which visually indistinguishable syllables may correspond to a large set of characters \cite{luo2025hard,zhao2020hearing}. This intrinsic ambiguity suggests that explicitly modeling an intermediate syllable-level representation is a more natural way to factorize uncertainty in Mandarin VSR.

Most existing VSR systems adopt an end-to-end modeling paradigm that directly maps video frames to word or character sequences \cite{fenghour2021deep} \cite{park2025swinlip} \cite{ma2022training} \cite{prajwal2022sub}. Recent studies have demonstrated that introducing explicit linguistic structure into the VSR pipeline can substantially improve recognition performance. \cite{yeo2024visual} \cite{liu2025leveraging} \cite{teng2025phoneme} \cite{jain2025hype}. In particular, the VALLR framework  \cite{thomas2025vallr} proposed by Thomas \emph{et al.} introduces a two-stage method for English lip-reading, where a neural network first predicts phoneme sequences from video, and a large language model subsequently decodes these phonemes into sentences. By explicitly modeling an intermediate phonetic representation, the framework allows the LLM to leverage linguistic priors to recover coherent text from noisy visual predictions, achieving strong performance with reduced amounts of visual training data.

However, unlike English, Mandarin does not use a phonemic alphabet, and its romanization system, Pinyin \cite{wikipedia:pinyin}, introduces additional complexity due to tone markings and many-to-one mappings between phonetic syllables and characters. Moreover, most existing LLM-based approaches implicitly assume relatively accurate phoneme or Pinyin inputs
. In practice, visual-only prediction of Pinyin remains error-prone, and naively feeding erroneous phonetic sequences into an LLM may amplify recognition errors rather than correct them.

Meanwhile, recent advances in automatic speech recognition (ASR) have shown that LLMs can be effectively used as post-processors for hypothesis rescoring and error correction \cite{tur2024progres} \cite{li2025large} \cite{ma2025asr}. By providing an LLM with N-best candidate transcriptions instead of a single hypothesis, the model can exploit linguistic context to resolve ambiguities and correct systematic recognition errors. Although this paradigm has shown promise in audio-based ASR, its application to visual speech recognition remains largely underexplored.

In this work, we propose VALLR-Pin, a Mandarin-oriented visual speech recognition framework that unifies visual modeling, phonetic representation, and language-level reasoning within a two-stage architecture. Instead of directly predicting characters from visual features, we reformulate Mandarin VSR as an uncertainty-factorized inference problem by jointly modeling syllable-level and character-level decoding processes, thereby explicitly decomposing visual uncertainty and linguistic ambiguity. On top of this factorized representation, we incorporate an LLM-based refinement module that operates under phonetic and candidate-level constraints to correct residual errors, rather than acting as an independent recognizer. Furthermore, to mitigate error propagation from imperfect phonetic predictions, we fine-tune the LLM using model-generated error patterns, enabling robust adaptation to the characteristic uncertainties of visual speech recognition.

\begin{figure*}[t]
  \centering
  \includegraphics[width=\textwidth]{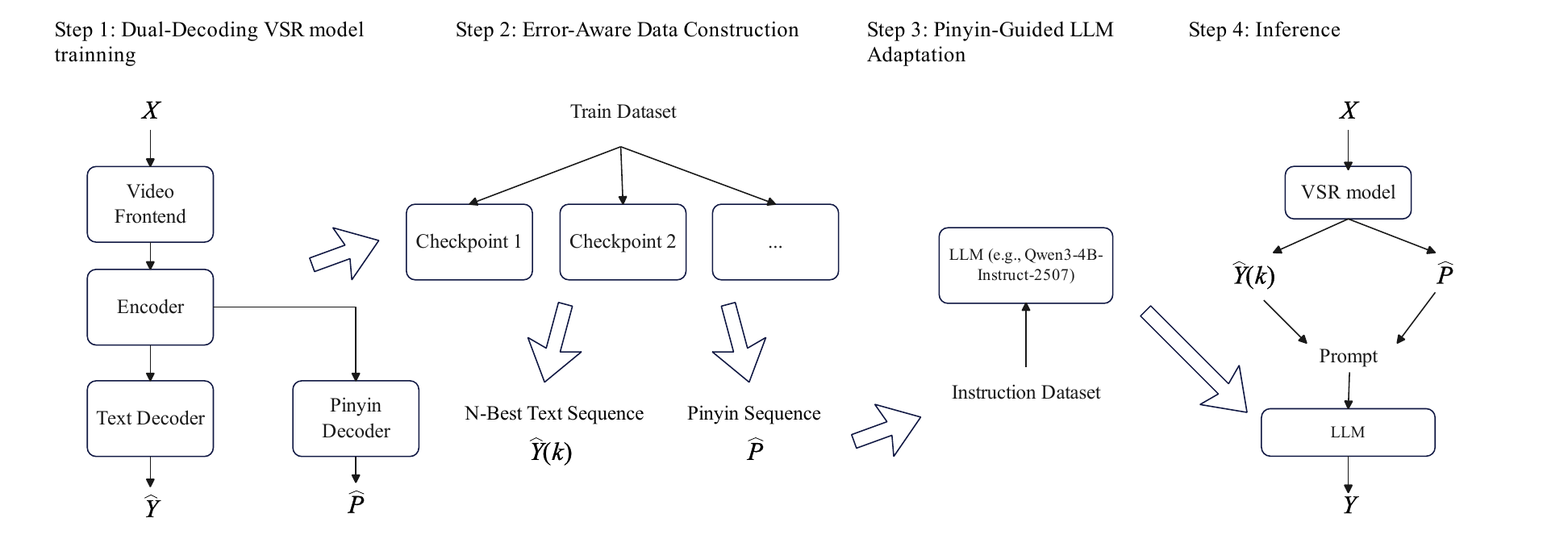}
  \caption{Overview of the proposed VALLR-Pin framework.
The system consists of a dual-decoder VSR model that predicts character and Pinyin sequences, followed by an LLM-based refinement module. Error-aware instruction data are constructed from model-generated hypotheses and used to adapt the LLM for Mandarin visual speech recognition.}
  \label{fig:framework}
\end{figure*}

\section{Proposal}
\label{sec:proposal}

In this section, we present the proposed \textbf{VALLR-Pin} framework, a two-stage visual speech recognition system for Mandarin that explicitly incorporates Pinyin as an intermediate representation and leverages a large language model for error-aware refinement.

\subsection{Problem Formulation}

Given a silent lip-reading video sequence
\begin{equation}
\mathbf{X} = \{ x_1, x_2, \dots, x_T \},
\end{equation}
where $x_t$ denotes the $t$-th video frame, the goal of visual speech recognition is to predict the corresponding Mandarin character sequence
\begin{equation}
\mathbf{Y} = \{ y_1, y_2, \dots, y_L \}.
\end{equation}

Conventional end-to-end VSR systems directly model the conditional distribution
\begin{equation}
P(\mathbf{Y} \mid \mathbf{X}),
\end{equation}
which is highly ambiguous in Mandarin due to extensive homophony and visually indistinguishable visemes.

To alleviate this issue, we introduce an explicit Pinyin sequence
\begin{equation}
\mathbf{P} = \{ p_1, p_2, \dots, p_M \},
\end{equation}
and decompose the overall inference process into two stages:
\begin{equation}
\mathbf{X}
\;\xrightarrow{\;\mathcal{F}_{\text{VSR}}\;}
(\hat{\mathbf{Y}}, \hat{\mathbf{P}})
\;\xrightarrow{\;\mathcal{F}_{\text{LLM}}\;}
\mathbf{Y},
\end{equation}
where $\mathcal{F}_{\text{VSR}}$ denotes a dual-decoder VSR model and $\mathcal{F}_{\text{LLM}}$ denotes an LLM-based refinement module.

\subsection{Stage I: Dual-Decoder Visual Speech Recognition}

\subsubsection{Visual Frontend and Encoder}

The input video sequence is first processed by a visual frontend to extract low-level spatiotemporal features:
\begin{equation}
\mathbf{F} = \text{Frontend}(\mathbf{X}),
\end{equation}
where the frontend consists of 3D convolutional layers followed by a ResNet-50 backbone \cite{theckedath2020detecting}.

The extracted features are then fed into a SANM encoder based on the Paraformer architecture \cite{gao22b_interspeech} \cite{shi2023seaco}, producing high-level visual-semantic representations:
\begin{equation}
\mathbf{H} = \text{Encoder}(\mathbf{F}), \quad \mathbf{H} \in \mathbb{R}^{T' \times d}.
\end{equation}

\subsubsection{Dual-Decoder Multi-Task Learning}

To jointly model phonetic and lexical information, we design a dual-decoder architecture that shares the same encoder output $\mathbf{H}$:

\begin{itemize}
    \item A \textbf{character decoder} based on SanmDecoder, modeling
    \begin{equation}
    P(\mathbf{Y} \mid \mathbf{H});
    \end{equation}
    \item A \textbf{Pinyin decoder} implemented with a lightweight Transformer decoder, modeling
    \begin{equation}
    P(\mathbf{P} \mid \mathbf{H}).
    \end{equation}
\end{itemize}

The Pinyin decoder predicts \emph{toneless} Pinyin units, resulting in a vocabulary of 397 modeling units, which simplifies the prediction space while preserving essential phonetic structure.

Both decoders are trained jointly using a combination of Connectionist Temporal Classification (CTC) \cite{graves2006connectionist} loss and cross-entropy (CE) loss. For the character decoder, the loss is defined as
\begin{equation}
\mathcal{L}_{\text{char}}
=
\lambda_{\text{ctc}} \mathcal{L}_{\text{ctc}}^{\text{char}}
+
(1 - \lambda_{\text{ctc}}) \mathcal{L}_{\text{ce}}^{\text{char}},
\end{equation}
and similarly for the Pinyin decoder:
\begin{equation}
\mathcal{L}_{\text{py}}
=
\lambda_{\text{ctc}} \mathcal{L}_{\text{ctc}}^{\text{py}}
+
(1 - \lambda_{\text{ctc}}) \mathcal{L}_{\text{ce}}^{\text{py}}.
\end{equation}

The overall training objective is a weighted sum of the two losses:
\begin{equation}
\mathcal{L}_{\text{total}}
=
\alpha \mathcal{L}_{\text{char}}
+
(1 - \alpha) \mathcal{L}_{\text{py}},
\end{equation}
where $\alpha$ is set to $0.5$ in all experiments. This multi-task formulation encourages the visual encoder to learn more robust and linguistically grounded representations.

\subsection{Stage II: LLM-Based Refinement with Pinyin Guidance}

\subsubsection{N-best Hypothesis Generation}

During inference, beam search is applied to the character decoder to generate a set of top-$K$ candidate transcriptions:
\begin{equation}
\{ \hat{\mathbf{Y}}^{(1)}, \hat{\mathbf{Y}}^{(2)}, \dots, \hat{\mathbf{Y}}^{(K)} \}.
\end{equation}
Meanwhile, the Pinyin decoder outputs the most probable Pinyin sequence:
\begin{equation}
\hat{\mathbf{P}}.
\end{equation}

These hypotheses provide complementary lexical and phonetic cues for downstream refinement.

\subsubsection{LLM Fine-Tuning Dataset Construction}

To adapt the LLM to the error characteristics of the VSR model, we construct a fine-tuning dataset using self-generated errors. Specifically, multiple checkpoints from early, middle, and late training stages of the VSR model are saved and used to decode the training set. These checkpoints yield predictions with varying character error rates (CERs), increasing data diversity.

Each training sample is formatted as an instruction-following instance:
\begin{equation}
P(\mathbf{Y} \mid \hat{\mathbf{P}}, \{ \hat{\mathbf{Y}}^{(k)} \}_{k=1}^{K}),
\end{equation}
where the input consists of a possibly erroneous Pinyin sequence $\hat{\mathbf{P}}$ and multiple candidate character hypotheses.

This formulation allows the LLM to better capture systematic error patterns produced by the VSR model.

\subsubsection{LLM Fine-Tuning with LoRA}

We adopt Low-Rank Adaptation (LoRA) \cite{hu2022lora} to efficiently fine-tune the LLM. Let $\theta$ denote the original LLM parameters. For selected weight matrices, LoRA introduces low-rank updates:
\begin{equation}
W = W_0 + \Delta W, \quad \Delta W = A B,
\end{equation}
where $A \in \mathbb{R}^{d \times r}$ and $B \in \mathbb{R}^{r \times d}$ with $r \ll d$.

The LLM is trained using an autoregressive language modeling objective:
\begin{equation}
\mathcal{L}_{\text{LLM}}
=
- \sum_{t}
\log P_{\theta}
\big(
y_t \mid y_{<t}, \hat{\mathbf{P}}, \{ \hat{\mathbf{Y}}^{(k)} \}
\big).
\end{equation}

Through fine-tuning, the LLM is guided to (1) map ambiguous Pinyin sequences to characters, (2) correct common visual confusion errors, and (3) exploit linguistic context to rescore and refine N-best hypotheses.

\subsection{Inference Summary}

The final prediction is obtained by selecting the most probable transcription under the refined LLM distribution:
\begin{equation}
\mathbf{Y}
=
\arg\max_{\mathbf{Y}}
P_{\text{LLM}}
\big(
\mathbf{Y}
\mid
\hat{\mathbf{P}}, \{ \hat{\mathbf{Y}}^{(k)} \}_{k=1}^{K}
\big).
\end{equation}

By explicitly integrating visual, phonetic, and linguistic information, VALLR-Pin provides a robust solution for Mandarin visual speech recognition.

\section{Experiments}
\label{sec:experiments}

This section evaluates the effectiveness of the proposed VALLR-Pin framework on Mandarin visual speech recognition. We conduct experiments on one public benchmark and one self-collected dataset with increased speaker diversity and visual complexity. Both quantitative and qualitative analyses are presented.

\subsection{Datasets}

\paragraph{CMLR}
This is a Chinese sentence-level lip-reading dataset, consisting of 11 speakers and 102,072 spoken sentences \cite{zhao2019cascade}. The model proposed by Pingchuan Ma et al. achieves a character error rate of 9.10 on this dataset \cite{ma2022visual}.

\paragraph{CNVSRC-Multi.Dev.}
We evaluate our method on the official multi-speaker evaluation set from the Chinese Continuous Visual Speech Recognition Challenge (CNVSRC) \cite{chen2024cnvsrc} \cite{liu2025cnvsrc}. The CNVSRC-Multi.Dev dataset consists of sentence-level silent lip-reading videos from 43 different speakers, including 23 speakers recorded in controlled indoor studio conditions and 20 speakers collected from Internet videos, providing a mix of controlled and real-world scenarios for multi-speaker visual speech recognition. 

For training, we adopt the fixed-track data specified in CNVSRC 2025 \cite{cnvsrc2025}. This includes four publicly released subsets: CN-CVS \cite{chen2023cn}, the core large-scale Mandarin visual-speech dataset, CNVSRC.Dev, CN-CVS2-P1, and CN-CVS3.

\paragraph{Self-Collected Dataset.}
To evaluate robustness under more realistic conditions, we construct a self-collected Mandarin lip-reading dataset featuring:
(i) a larger number of speakers,
(ii) diverse recording environments,
and (iii) increased variability in pose, illumination, and speaking style.
Compared with CNVSRC-Multi, this dataset exhibits higher visual ambiguity and stronger domain shift.
We follow a speaker-dependent split and evaluate performance on a held-out unseen-speaker test set.

\subsection{Evaluation Metric}

We adopt CER as the evaluation metric:
\begin{equation}
\text{CER} = \frac{S + D + I}{N},
\end{equation}
where $S$, $D$, and $I$ denote the numbers of substitutions, deletions, and insertions, respectively, and $N$ is the number of characters in the ground-truth transcription.

\subsection{Experimental Setup}

\paragraph{VSR Models.}
We build a char-only baseline model by integrating the Paraformer structure into the CNVSRC 2025 baseline model. All VSR models share the same visual frontend consisting of 3D convolutional layers followed by a ResNet-50 backbone. The encoder is based on a SANM Encoder.
For decoding, the character-only baseline employs a SANM decoder; on this basis, our improved model introduces an additional Transformer decoder-based Pinyin decoder, enabling it to simultaneously predict Mandarin characters and toneless Pinyin sequences.

\paragraph{LLM Refinement.}
For the second-stage refinement, we fine-tune a large language model, Qwen3-4B-Instruct-2507 \cite{qwen3technicalreport} \footnote{https://modelscope.cn/models/Qwen/Qwen3-4B-Instruct-2507}, using the ms-swift framework \cite{zhao2024swiftascalablelightweightinfrastructure}.
We adopt LoRA and accelerate training with DeepSpeed \cite{rasley2020deepspeed}.
The main hyperparameters are:
\texttt{learning\_rate} $= 1\mathrm{e}{-4}$,
\texttt{lora\_rank} $= 8$,
and \texttt{lora\_alpha} $= 32$.
The LLM is fine-tuned on instruction-following data constructed from model-generated errors using multiple VSR checkpoints.
During inference, the LLM takes the predicted Pinyin sequence and an N-best list of character hypotheses as input.

\subsection{Comparison with Baselines}

Table~\ref{tab:main_results} compares the proposed method with baseline systems on both datasets.

\begin{table}[t]
\centering
\caption{CER (\%) comparison on CNVSRC-Multi.Dev and the self-collected dataset. Lower is better.}
\label{tab:main_results}
\begin{tabular}{lccc}
\hline
\textbf{Method} & \textbf{Setting} & \textbf{CNVSRC-Multi.Dev} & \textbf{Self-Collected} \\
\hline
2023 Baseline & S1 & 58.77 & - \\
2024 Baseline & S2 & 52.42 & - \\
2025 Baseline & S3 & 31.91 & 40.21 \\
Char-only Baseline & S3 & 30.87 & 38.60 \\
\textbf{VALLR-Pin} & S3 & \textbf{28.39} & \textbf{35.49} \\
\textbf{VALLR-Pin} & open & 24.10 & 32.22 \\
\hline
\end{tabular}
\parbox{\linewidth}{
  \vspace{2mm}  
  \scriptsize
  \textbf{Settings:} 
  S1=CN-CVS+CNVSRC.Dev; 
  S2=S1+CN-CVS2-P1; 
  S3=S2+CN-CVS3.
}
\end{table}

\begin{table}[t]
\centering
\caption{CER (\%) comparison on CMLR. Lower is better.}
\label{tab:cmlr_results}
\begin{tabular}{lc}
\hline
\textbf{Model} & \textbf{CER} \\
\hline
Ma's model & 9.10 \\
+ Pinyin-Guided Refinement & 7.89 \\
\hline
\end{tabular}
\end{table}

The char-only baseline achieves a lower CER than the CNVSRC 2025 competition baseline. Additionally, the proposed VALLR-Pin consistently outperforms all baselines.
The improvement is more pronounced on the self-collected dataset, indicating stronger robustness under complex visual conditions.

Additionally, we conduct a comparative experiment with Pingchuan Ma's model on the CMLR dataset; integrating the dual-decoding structure and two-stage LLM Refinement into their model leads to a further reduction in CER, with results presented in Table~\ref{tab:cmlr_results}.

\subsection{Ablation Studies}

\paragraph{Effect of Dual-Decoder Architecture.}
Table~\ref{tab:ablation_decoder} evaluates the impact of introducing the Pinyin decoder.

\begin{table}[t]
\centering
\caption{Ablation on dual-decoder design (CER \%).}
\label{tab:ablation_decoder}
\begin{tabular}{lc}
\hline
\textbf{Model} & \textbf{CER} \\
\hline
Char-only Baseline & 38.60 \\
  + Pinyin Decoder & 37.23 \\
\hline
\end{tabular}
\end{table}

The results confirm that jointly predicting Pinyin provides effective phonetic constraints and improves visual-semantic representations.

\paragraph{Effect of LLM Fine-Tuning.}
Table~\ref{tab:ablation_llm} compares different LLM configurations.

\begin{table}[t]
\centering
\caption{Ablation on LLM refinement (CER \%).}
\label{tab:ablation_llm}
\begin{tabular}{lc}
\hline
\textbf{LLM Setting} & \textbf{CER} \\
\hline
No LLM & 37.23 \\
  + Zero-shot LLM & 37.86 \\
  + LLM Refinement & \textbf{35.49} \\
\hline
\end{tabular}
\end{table}

Fine-tuning the LLM on model-generated errors yields substantial gains over zero-shot usage, demonstrating the importance of error-aware adaptation.

\begin{CJK}{UTF8}{gbsn}
\begin{table}[!ht]
\centering
\caption{Qualitative examples of transcription refinement.}
\label{tab:qualitative}
\begin{tabular}{p{2.4cm} p{2.4cm} p{2.5cm}}
\hline
\textbf{Ground Truth} & \textbf{Stage I Output} & \textbf{After LLM Refinement} \\
\hline
我想去银行 & 我想去银航 & 我想去银行 \\ 
\textit{I want to go to the bank} & \textit{I want to go to the silver flight} & \textit{I want to go to the bank} \\
\hline
今天可能会下雨 & 今天可能会夏雨 & 今天可能会下雨 \\ 
\textit{It may rain today} & \textit{It may summer rain today}
 & \textit{It may rain today} \\
\hline
他正在开会讨论 & 他正在开会导论 & 他正在开会讨论 \\ 
\textit{He is in a meeting discussing}
 & \textit{He is in a meeting introducing}
 & \textit{He is in a meeting discussing} \\
\hline
\end{tabular}
\end{table}
\end{CJK}

\subsection{Qualitative Analysis}

Table~\ref{tab:qualitative} presents representative examples illustrating the effect of LLM-based refinement.
The LLM effectively resolves homophone errors and visually confusable characters by leveraging Pinyin guidance and contextual information.

\section{Conclusion}
\label{sec:conclusion}
We presented VALLR-Pin, an uncertainty-aware framework for Mandarin visual speech recognition that explicitly factorizes visual ambiguity through an intermediate syllable-level representation. By decomposing Mandarin VSR into coupled syllable-level and character-level inference processes, the proposed approach addresses the inherently ill-posed nature of direct character prediction from visual input. On top of this factorized representation, a lightweight LLM-based refinement module is employed as a complementary constraint mechanism to correct residual linguistic ambiguities, rather than acting as an independent recognizer. Experimental results on the CNVSRC benchmark demonstrate consistent improvements under multi-speaker conditions, indicating that uncertainty-aware modeling combined with constrained language-level refinement provides an effective and modular enhancement to visual speech recognition systems.

\end{document}